# Stiffness modeling of non-perfect parallel manipulators

Klimchik A., Pashkevich A. and Chablat D.

*Abstract*— The paper focuses on the stiffness modeling of parallel manipulators composed of non-perfect serial chains, whose geometrical parameters differ from the nominal ones. In these manipulators, there usually exist essential internal forces/torques that considerably affect the stiffness properties and also change the end-effector location. These internal loadings are caused by elastic deformations of the manipulator elements during assembling, while the geometrical errors in the chains are compensated for by applying appropriate forces. For this type of manipulators, a non-linear stiffness modeling technique is proposed that allows us to take into account inaccuracy in the chains and to aggregate their stiffness models for the case of both small and large deflections. Advantages of the developed technique and its ability to compute and compensate for the compliance errors caused by different factors are illustrated by an example that deals with parallel manipulators of the Orthoglide family.

## I. INTRODUCTION

IN modern industrial robotics, stiffness becomes one of the most important performance measures that defines potential accuracy of the manipulator. This problem has been the focus of numerous works [1-5], where different solutions for serial and parallel manipulators have been proposed assuming that the manipulator geometry perfectly corresponds to the nominal one. However in practice, parallel manipulators are usually composed of non-perfect serial chains, whose geometrical parameters differ from the nominal values. It is evident that these manufacturing errors may generate essential internal forces and have effects on the manipulator stiffness behavior. However, this problem has attracted very limited attention in robotics.

In general, there exist several stiffness modeling methods, which were analyzed in details in our previous works [6-8].

For industrial applications, the most popular technique is based on the Virtual Joint Modeling (VJM) approach that was first introduced in robotics by Salisbury and Gosselin [9-10] and has been further developed by a number of authors [11-14]. It extends the conventional rigid-body model of the robotic manipulator by adding virtual springs that take into account elastostatic properties of links and joints. In the first works, it was explicitly assumed that the main sources of elasticity are concentrated in actuated joints [9-10]. Correspondingly, the links were assumed to be rigid and the VJM model included one-dimensional springs only. Recent modifications of this approach describes elastostatic properties of links using $6 \times 6$ non-diagonal stiffness matrices [7] that are computed taking into account the real shape of complex links [15]. Using this approach it is possible to obtain a rather general non-linear stiffness model for a serial chain [7] and to compute the Cartesian stiffness matrix even for singular configurations.

For parallel robots [16], the VJM technique can be applied either in a straightforward way (by considering the static equilibrium equations for all chains simultaneously [10][17][18]) or by decomposing the manipulator into a set of separate serial chains, obtaining the stiffness models for all of them and further aggregation of these models in a united one corresponding to the parallel manipulator. It is obvious that the first approach, which incorporates a solution of high order non-linear matrix equations [19], is rather tedious to be applied to real life industrial problems. In contrast, the second approach relies on relatively simple techniques that are well developed for serial manipulators. The latter was partially implemented in [6][20], where the manipulator structure was assumed to be strictly parallel (i.e. without internal loops) and the kinematic chains where assembled in the same end-point. Under this assumption, the stiffness matrix of the parallel manipulator can be computed via simple summation of the chain stiffness matrices. However, in more general (and practically important) cases where the kinematic chains are connected to different points of the end-platform, this technique cannot be applied directly.

Another limitation of existing results in this area is related to the assumption that the assembling does not produce any internal forces/torques. But in practice, numerous errors are accumulated in serial chains [21] and they cause non-negligible internal forces in manipulator joints (even if the external force applied to the end-effector is equal to zero). Furthermore, the kinematic chains of the robotic manipula-

Manuscript received March 7, 2011. The work presented in this paper was partially funded by the Region "Pays de la Loire", France (Project RoboComposite) and by the The French National Research Agency (Project ANR-2010-SEGI-003-02-COROUSSO).

A. Klimchik is with Ecole des Mines de Nantes, 4 rue Alfred-Kastler, Nantes 44307, France and with Institut de Recherches en Communications et Cybernétique de Nantes, 44321 Nantes, France (phone: Tel.+33-251-85-83-17; fax.+33-251-85-83-49; e-mail: alexandr.klimchik@mines-nantes.fr).

A. Pashkevich is with Ecole des Mines de Nantes and with Institut de Recherches en Communications et Cybernétique de Nantes (e-mail: anatol.pashkevich@mines-nantes.fr).

D. Chablat is with Institut de Recherches en Communications et Cybernétique de Nantes (e-mail: damien.chablat@irccyn.ec-nantes.fr).

tors may include some additional elastic elements in the actuated or/and passive joints that are intended to increase the robot positioning accuracy or to improve the stiffness properties in certain workspace areas. For example, to eliminate the backlash, the gear trains may include spring-loaded scissor elements that generate the internal forces, which must also be integrated in the stiffness model [22]. Similar forces may also arise in the parallel manipulators with antagonistic actuating [18]. Other examples include parallel manipulators with springs interposed in the passive joints in order to improve stiffness in the singularity neighborhood.

As follows from relevant studies performed by the authors, the internal forces may essentially influence the manipulator behavior (modify the stiffness matrix, change the end-effector location, etc.) and should be obviously taken into account in the stiffness model. However, most existing works ignore this issue.

Thus, this paper focuses on the stiffness modeling of parallel manipulators with non-perfect serial chains. To address this problem, the remainder of the paper is organized as follows: Section II proposes stiffness modeling background, Section III deals with aggregation of stiffness models without loading, Section IV extends aggregation technique for the case of loaded mode, Section V illustrates the developed technique by the example of Orthoglide manipulator and, finally, Section VI summarizes the main contributions.

## II. STIFFNESS MODELING BACKGROUND

The stiffness modeling technique being developed in this work is based on our previous results [6][7], that deal with perfect manipulators. Let us present them briefly.

For the considered manipulators, if the external loading is equal to zero, all kinematic chains can be aligned and matched in the same target point $\mathbf{t}_0$. In the neighborhood of this point, for each $i$th kinematic chain the desired stiffness model is defined by the non-linear force-deflection relation

$$\mathbf{F}_i = f_i(\mathbf{t} \mid \mathbf{t}_0) \tag{1}$$

where $\mathbf{t}$ denotes the end-effector location and $\mathbf{F}_i$ is the corresponding external loading applied to the chain end-point. To obtain the function $f_i(.)$, the following iterative procedure can be used

$$\begin{bmatrix} \mathbf{F}'_i \\ \mathbf{q}'_i \\ \boldsymbol{\theta}'_i \end{bmatrix} = \begin{bmatrix} \mathbf{0} & \mathbf{J}_{qi} & \mathbf{J}_{\theta i} \\ \mathbf{J}_{qi}^T & \mathbf{0} & \mathbf{0} \\ \mathbf{J}_{\theta i}^T & \mathbf{0} & -\mathbf{K}_{\theta i} \end{bmatrix}^{-1} \begin{bmatrix} \mathbf{t} - \mathbf{g}_i + \mathbf{J}_{qi} \mathbf{q}_i + \mathbf{J}_{\theta i} \boldsymbol{\theta}_i \\ \mathbf{0} \\ -\mathbf{K}_{\theta i} \boldsymbol{\theta}_{0i} \end{bmatrix} \tag{2}$$

where the subscript "$i$" denotes the serial chain number, the prime corresponds to the next iteration, $(\mathbf{q}_i, \boldsymbol{\theta}_i)$ defines the chain configuration that depends on the passive and virtual joint coordinates $\mathbf{q}_i$ and $\boldsymbol{\theta}_i$ respectively, $\mathbf{J}_q(\mathbf{q}, \boldsymbol{\theta})$ and $\mathbf{J}_\theta(\mathbf{q}, \boldsymbol{\theta})$ are corresponding Jacobian matrices computed for current configuration, $\boldsymbol{\theta}_0$ is preloading in the virtual joints, matrix $\mathbf{K}_\theta$ describes the joints stiffness properties, function $\mathbf{t} = \mathbf{g}_i(\mathbf{q}_i, \boldsymbol{\theta}_i)$ defines the chain geometry.

After linearization, for each given $\mathbf{t}$, the Cartesian stiffness matrices $\mathbf{K}_C^{(i)}$ of all chains can be computed as

$$\mathbf{K}_C = \mathbf{K}_C^{0(F)} - \mathbf{K}_C^{0(F)} \cdot \left( \mathbf{J}_q + \mathbf{J}_\theta \cdot \mathbf{k}_\theta^F \cdot \mathbf{H}_{\theta q}^F \right) \mathbf{K}_{Cq} \tag{3}$$

where the first term $\mathbf{K}_C^{0(F)} = (\mathbf{J}_\theta \mathbf{k}_\theta^F \mathbf{J}_\theta^T)^{-1}$ corresponds to the classical formula defining the stiffness of the kinematic chain without passive joints in the loaded mode [14] and the second term takes into account the influence of the passive joints. Besides, the stiffness matrix $\mathbf{K}_{Cq}$ (defining a linear mapping of the end-point displacement $\delta \mathbf{t}$ to the deflections in the passive joints $\delta \mathbf{q}$) can be computed as

$$\mathbf{K}_{Cq} = -\left( \mathbf{H}_{qq}^F + \mathbf{H}_{q\theta}^F \mathbf{k}_\theta^F \mathbf{H}_{\theta q}^F - \left( \mathbf{J}_q^T + \mathbf{H}_{q\theta}^F \mathbf{k}_\theta^F \mathbf{J}_\theta^T \right) \cdot \\ \cdot \mathbf{K}_C^{0(F)} \left( \mathbf{J}_q + \mathbf{J}_\theta \mathbf{k}_\theta^F \mathbf{H}_{\theta q}^F \right) \right)^{-1} \left( \mathbf{J}_q^T + \mathbf{H}_{q\theta}^F \mathbf{k}_\theta^F \mathbf{J}_\theta^T \right) \mathbf{K}_C^{0(F)} \tag{4}$$

Here $\mathbf{k}_\theta^F = (\mathbf{K}_\theta - \mathbf{H}_{\theta\theta}^F)^{-1}$ and $\mathbf{H}_{v_1 v_2}^F = \partial^2 \left( \mathbf{g}^T \mathbf{F} \right) / \partial \mathbf{v}_1 \partial \mathbf{v}_2$ are the Hessian matrices with respect to combination of the passive and virtual joint coordinates $(\mathbf{q}, \mathbf{q})$, $(\mathbf{q}, \boldsymbol{\theta})$, $(\boldsymbol{\theta}, \mathbf{q})$, $(\boldsymbol{\theta}, \boldsymbol{\theta})$.

In addition, linearization provides the matrix $\mathbf{K}_{C\theta}$ that defines linear mappings of the end-point displacement $\delta \mathbf{t}$ to the virtual joint coordinates $\delta \boldsymbol{\theta}$

$$\mathbf{K}_{C\theta} = \mathbf{k}_\theta^F \mathbf{J}_\theta^T \mathbf{K}_C + \mathbf{k}_\theta^F \mathbf{H}_{\theta q}^F \mathbf{K}_{Cq} \tag{5}$$

that is computed using the same intermediate variables.
This approach allows us to obtain the non-linear force-deflection relation for each serial chain as well as to compute the Cartesian stiffness matrices for any given target point $\mathbf{t}_0$ and given the end-point location $\mathbf{t}$. However, it cannot be applied directly for parallel manipulators with non-perfect serial chains because it is implicitly assumed here that assembling in the point $\mathbf{t}_0$ does not require any forces applied to the chain end-point, i.e. $f_i(\mathbf{t}_0 \mid \mathbf{t}_0) = \mathbf{0}$. Thus, a dedicated technique, which is considered in this paper, is required.

## III. STIFFNESS MODELS AGGREGATION FOR SMALL LOADING

If the external loading applied to the mobile platform of the parallel manipulator is small enough, and a linearization-based approach is reasonable. Below, is briefly presented for the case of perfect kinematic chains [8] and then developed in more details for non-perfect chains.

### A. Stiffness model aggregation for perfect chains

In this case, it is assumed that all the chains can be assembled in the target point $\mathbf{t}_0$ without any external loading that may be expressed as $f_i(\mathbf{t}_0 \mid \mathbf{t}_0) = \mathbf{0}$. So, for the parallel manipulator, the desired force-deflection relation can be written as

$$\mathbf{F} = \mathbf{K}_C \cdot (\mathbf{t} - \mathbf{t}_0) \quad (6)$$

where the total Cartesian stiffness matrix $\mathbf{K}_C$ is computed using the chain aggregation formula

$$\mathbf{K}_C = \sum_{i=1}^{m} \mathbf{J}_v^{(i)-T} \mathbf{K}_C^{(i)} \mathbf{J}_v^{(i)-1} \quad (7)$$

which integrates the Cartesian stiffness matrices $\mathbf{K}_C^{(i)}$ of all $m$ chains taking into account the difference between the reference point of the end-platform and the end-points of the chains (where the chains are connected to the mobile platform, Fig. 1). The latter is expressed via the Jacobians $\mathbf{J}_v^{(i)}$

$$\mathbf{J}_v^{(i)-1} = \begin{bmatrix} \mathbf{I}_3 & -(\mathbf{v}_i \times) \\ \mathbf{0} & \mathbf{I}_3 \end{bmatrix}_{6\times 6} \quad (8)$$

where $\mathbf{I}_3$ is $3\times 3$ identity matrix, $(\mathbf{v}\times)$ is a skew-symmetric matrix corresponding to the vector $\mathbf{v}$, vector $\mathbf{v}_i$ defines distance from the leg end-point to the end-effector reference point (see Fig.1).

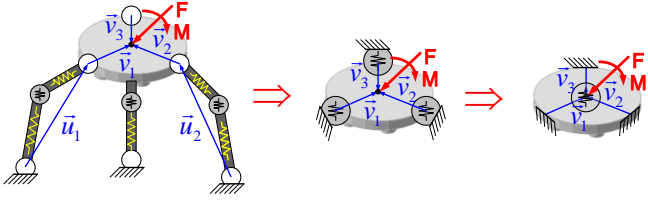

Fig. 1. Transformation of VJM models of typical parallel manipulator

Further, linear relations between the end-platform displacement $\mathbf{t} - \mathbf{t}_0$ and variations $\delta \mathbf{q}_i$, $\delta \mathbf{\theta}_i$ in the joint coordinates of the chains may be presented as

$$\delta \mathbf{q}_i = \mathbf{K}_{Cq}^{(i)} \cdot \mathbf{J}_v^{(i)-1} \cdot (\mathbf{t} - \mathbf{t}_0); \quad \delta \mathbf{\theta}_i = \mathbf{K}_{C\theta}^{(i)} \cdot \mathbf{J}_v^{(i)-1} \cdot (\mathbf{t} - \mathbf{t}_0) \quad (9)$$

where, the joint sensitivity matrices $\mathbf{K}_{Cq}^{(i)}$, $\mathbf{K}_{C\theta}^{(i)}$ are computed from (4) and (5) assuming that $\mathbf{F} = \mathbf{0}$ and by neglecting all Hessians (here, the matrices $\mathbf{K}_{Cq}^{(i)}$, $\mathbf{K}_{C\theta}^{(i)}$ are expressed with respect to the chain end-points).

### B. Stiffness model aggregation for non-perfect chains

If the kinematic chains are non-perfect, the corresponding force-deflection relation (1) is shifted with respect to the point $\mathbf{t}_0$, i.e. $f_i(\mathbf{t}_0 | \mathbf{t}_0) \neq 0$. So, the manipulator assembling in this point requires application of some non-zero forces $\mathbf{F}_i$ that generally do not compensate each other. Correspondingly, the end-platform location differs from $\mathbf{t}_0$ if the external force applied to the end-platform is equal to zero. Let us denote this difference as $\Delta \mathbf{t}$ and revise the above matrix equations (6)-(9) assuming that, without the external loading, the chain end-point is shifted by $\mathbf{\varepsilon}_i$ with respect to $\mathbf{t}_0$ (it can be also expressed as $f_i(\mathbf{t}_0 + \mathbf{\varepsilon}_i | \mathbf{t}_0) = 0$).

Using these notations, the desired stiffness model can be described by the following expressions

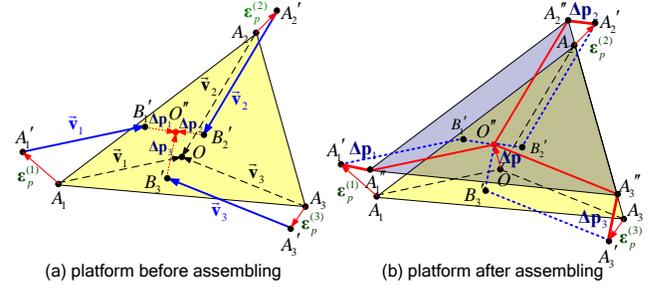

(a) platform before assembling     (b) platform after assembling

Fig. 2 Transformation of characteristic points of serial chains in assembling of non-perfect parallel manipulator; ($A_i$, $A_i'$ - end-point locations of serial chain before assembling for perfect and non-perfect manipulators respectively, $A_i''$ - end-point location of serial chain after assembling for non-perfect manipulator)

$$\mathbf{F} = \mathbf{K}_C \cdot (\mathbf{t} - \mathbf{t}_0 + \Delta \mathbf{t}) \quad (10)$$

and

$$\delta \mathbf{q}_i = \mathbf{K}_{Cq}^{(i)} \cdot \mathbf{J}_v^{(i)-1} \cdot (\mathbf{t} - \mathbf{t}_0 + \Delta \mathbf{t}_i);$$
$$\delta \mathbf{\theta}_i = \mathbf{K}_{C\theta}^{(i)} \cdot \mathbf{J}_v^{(i)-1} \cdot (\mathbf{t} - \mathbf{t}_0 + \Delta \mathbf{t}_i) \quad (11)$$

where the vectors $\Delta \mathbf{t}$, $\Delta \mathbf{t}_i$ should be computed using the assumptions presented above.

To find these additional parameters of the stiffness model, let us apply the energy based approach which allows us to express the potential energy of the parallel manipulator as

$$E(\Delta \mathbf{t}) = \frac{1}{2} \sum_{i=1}^{m} \left( (\mathbf{\varepsilon}_i - \Delta \mathbf{t})^T \cdot \mathbf{K}_C^{(i)} \cdot (\mathbf{\varepsilon}_i - \Delta \mathbf{t}) \right) \quad (12)$$

where $\Delta \mathbf{t} = (\Delta \mathbf{p}, \Delta \mathbf{\varphi})$ is displacement (position and orientation) of the reference point, $\mathbf{K}_C^{(i)}$ is the Cartesian stiffness matrix of the $i$th chain, and $\mathbf{\varepsilon}_i$ is the vector that integrates influence of all geometrical errors at the reference point.

It is obvious that after assembling, the considered mechanical system will occupy the most advantageous configuration with respect to the potential energy. It means that the desired vector $\Delta \mathbf{t}$ can be found from the expression $\partial E / \partial \Delta \mathbf{t} = 0$, which allows us to evaluate the end-platform deflection

$$\Delta \mathbf{t} = \left( \sum_{i=1}^{m} \mathbf{K}_C^{(i)} \right)^{-1} \cdot \sum_{i=1}^{m} \left( \mathbf{K}_C^{(i)} \cdot \mathbf{\varepsilon}_i \right) \quad (13)$$

and the end-platform location after assembling $\mathbf{t}' = \mathbf{t}^0 + \Delta \mathbf{t}$. Hence, for each separate kinematic chain, the end-frame deflections $\Delta \mathbf{t}_i = \Delta \mathbf{t} - \mathbf{\varepsilon}_i$ due to assembling is expressed as

$$\Delta \mathbf{t}_i = \left( \sum_{i=1}^{m} \mathbf{K}_C^{(i)} \right)^{-1} \cdot \sum_{i=1}^{m} \left( \mathbf{K}_C^{(i)} \cdot \mathbf{\varepsilon}_i \right) - \mathbf{\varepsilon}_i \quad (14)$$

and the corresponding loading applied to the end-point (due to interaction with other non-perfect chains) is $\mathbf{F}_i = \mathbf{K}_C^{(i)} \cdot \Delta \mathbf{t}_i$. Accordingly, the loadings in the virtual joints $\mathbf{\tau}_\theta^{(i)} = \mathbf{J}_\theta^{(i)T} \cdot \mathbf{F}_i$ can be computed as

$$\boldsymbol{\tau}_{\theta}^{(i)} = \mathbf{J}_{\theta}^{(i)\mathrm{T}} \cdot \mathbf{K}_{\mathrm{C}}^{(i)} \cdot \Delta \mathbf{t}_i \qquad (15)$$

It is worth mentioning that here $\sum_{i=1}^{m} \mathbf{F}_i = \mathbf{0}$, since there is no external loading applied to the platform reference point after the assembling. Besides, it is possible to compute relevant deflections of the virtual and passive joint coordinates of the chains caused by the assembling

$$\boldsymbol{\theta}_i = \mathbf{K}_{\mathrm{C}\theta}^{(i)} \cdot \Delta \mathbf{t}_i; \qquad \Delta \mathbf{q}_i = \mathbf{K}_{\mathrm{Cq}}^{(i)} \cdot \Delta \mathbf{t}_i \qquad (16)$$

Thus, the above expressions allow us to evaluate the end-platform deflection and internal forces/torques caused by assembling of kinematic chains with geometrical errors. However, the total manipulator Cartesian stiffness matrix $\mathbf{K}_\mathrm{C}$ is the same as in Section III.A, since the geometrical errors are assumed to be small enough.

## IV. STIFFNESS MODELS AGGREGATION FOR HIGH LOADING

If the external loading is rather high, the manipulator stiffness model is essentially non-linear and the above presented techniques cannot be applied directly. However, some basic ideas from Section III can also be adopted here.

### A. Stiffness model of parallel manipulator

Let us focus first on the aggregation of stiffness models of separate serial chains into the stiffness model of the whole parallel manipulator in the loaded mode. To solve this problem, it is necessary to obtain the non-linear force-deflection relation, which takes into account elastostatic properties of all kinematic chains, and to compute corresponding Cartesian stiffness matrix.

Let us assume that, for the perfect kinematic chains, their end-points may be aligned and matched in the same target point $\mathbf{t}_0$, which corresponds to the desired end-platform location. This point is assumed to be known and allows us to compute (from the inverse kinematic model) the actuator and passive joint coordinates defining nominal configurations of the chains $(\mathbf{q}_{0i}, \boldsymbol{\theta}_{0i})$. It is also assumed that the stiffness models of all kinematic chains have been already obtained using techniques proposed in Sections II and are presented in the form of non-linear force-deflection relations $\mathbf{F}_i = f_i(\mathbf{t} \mid \mathbf{t}_0)$ corresponding to the target point $\mathbf{t}_0$.

It is evident that the external loading $\mathbf{F}$ changes the end-platform location $\mathbf{t}_0$, hence it is reasonable to consider the set of locations $\mathbf{t}$ in the neighborhood of target one. Under the above assumptions, for any given point $\mathbf{t}$ (from neighborhood of $\mathbf{t}_0$), it is possible to compute both the forces $\mathbf{F}_i$ and corresponding equilibrium configurations $(\mathbf{q}_i, \boldsymbol{\theta}_i)$. Then, in accordance with the superposition principle, the desired non-linear force-deflection relation for the whole parallel manipulator can be found by straightforward summation of all partial forces $\mathbf{F}_i$, i.e.

$$\mathbf{F} = \sum_{i=1}^{m} f_i(\mathbf{t} \mid \mathbf{t}_0) \qquad (17)$$

where $\mathbf{F}$ denotes the total external loading applied to the end-platform. Corresponding curves can be obtained by multiple repetition of the above described procedures for different values of the end-platform location $\mathbf{t}$.

Furthermore, for each given $\mathbf{t}$, the stiffness matrices $\mathbf{K}_\mathrm{C}^{(i)}$ of all kinematic chains can be computed using expression (3) So, the Cartesian stiffness matrix $\mathbf{K}_\mathrm{C}$ of the whole parallel manipulator is the sum

$$\mathbf{K}_\mathrm{C} = \sum_{i=1}^{m} \mathbf{K}_\mathrm{C}^{(i)} \qquad (18)$$

However, the matrices $\mathbf{K}_{\mathrm{Cq}}^{(i)}$ and $\mathbf{K}_{\mathrm{C}\theta}^{(i)}$ defining the "sensitivity" of the chain joint coordinates $(\mathbf{q}_i, \boldsymbol{\theta}_i)$ to the end-platform displacement cannot be aggregated in this way. They should be computed separately to evaluate forces/torques in the joints/links $\boldsymbol{\tau}_{\theta}^{(i)} = \mathbf{J}_{\theta}^{(i)\mathrm{T}} \cdot \mathbf{F}_i$, where $\mathbf{J}_{\theta}^{(i)}$ is Jacobian matrix of $i$-th kinematic chain with respect to virtual joint coordinates (see equations (4) and (5)).

It is worth mentioning that it was implicitly assumed above that the manipulator assembling is equivalent to the aligning and matching of the chain end-frames. To deal with a more general case, when the chains are connected to the different points of the platform, it is necessary to slightly modify the chain geometrical models and to re-compute the forces/torques and the stiffness matrices by adding a virtual rigid link connecting the end-point of the chain and the reference point of the platform. After relevant transformations, the presented technique above can be applied straightforwardly, using equations (7) and (9).

Besides, in contrast to Section III, here there are no evident differences in stiffness models aggregation of perfect and non-perfect kinematic chains. In the last case, the chain geometrical errors $\boldsymbol{\varepsilon}_i$ are implicitly included in the force-deflection relations $\mathbf{F}_i = f_i(\mathbf{t} \mid \mathbf{t}_0)$ in such a way that $f_i(\mathbf{t}_0 + \boldsymbol{\varepsilon}_i \mid \mathbf{t}_0) = \mathbf{0}$. As a result, the end-platform cannot be located in the target point $\mathbf{t}_0$ without external loading, i.e. $\sum_{i=1}^{m} f_i(\mathbf{t}_0 \mid \mathbf{t}_0) \neq \mathbf{0}$. Moreover, without external loading, the end-platform location $\mathbf{t}_\varepsilon$ differs from the target one $\mathbf{t}_0$ by the vector $\Delta \mathbf{t}$ that can be computed as

$$\Delta \mathbf{t} = \arg \left\{ \sum_{i=1}^{m} f_i\left(\mathbf{t}_0 + \Delta \mathbf{t} \mid \mathbf{t}_0\right) = \mathbf{0} \right\} \qquad (19)$$

Relevant numerical routines is presented below.

Corresponding internal forces $\mathbf{F}_i$ defining the chain internal loadings due to the geometrical errors can be computed by simple substitution $\mathbf{t}_0 + \Delta \mathbf{t}$ to the partial force deflection relations $\mathbf{F}_i = f_i(\mathbf{t}_0 + \Delta \mathbf{t} \mid \mathbf{t}_0)\big|$. It is obvious that the sum of the $\mathbf{F}_i$ is equal to zero but they produce forces/torques in the links and joints if the parallel manipulator is over-constrained.

Hence, the developed aggregation technique allows us to obtain the non-linear force-deflection relation for a parallel manipulator in the loaded mode as well as to compute Carte-sian stiffness matrices for any given target point $t_0$ and given set of the end-point locations $\{t\}$. This technique is summarized in Fig. 3.

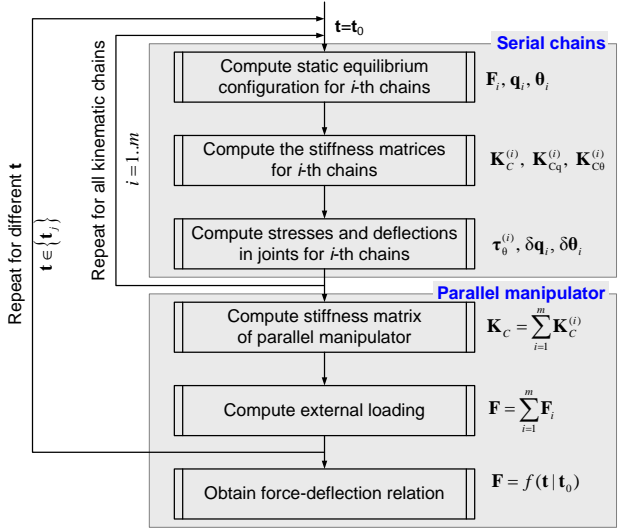

**Fig. 3** Aggregation of serial chains stiffness models technique

### B. Compliance model of parallel manipulator

The non-linear force-deflection relation (17) allows us to evaluate the external force/torque $F$ required to locate the manipulator at any given point $t$ (assuming that the actuated coordinates are computed for the end-platform location $t_0$ corresponding to the unloaded configuration). However in practice, it is often necessary to determine the end platform resistance to the external loading, i.e. to compute the deflection caused by the force $F$ applied to the end-platform. The desired value can be found from the non-linear compliance model that in a general case is expressed as

$$t = f^{-1}(F \mid t_0) \tag{20}$$

and is defined by the inverse function $f^{-1}(...)$ which for parallel manipulators usually exists (due to the overconstrained structure). In contrast, for serial chains with passive joints, the function $f^{-1}(...)$ cannot be computed since the corresponding Cartesian stiffness matrix $K_C^{(i)}$ is singular.

It is obvious that in a general case, the function $f^{-1}(...)$ cannot be expressed analytically. Hence, a dedicated iterative procedure, which is able to solve the non-linear equation (17) for $t$ (assuming that $F$ is given) is required. It is proposed here to apply a modification of Newton-Raphson technique which iteratively updates the desired value $t$ in accordance with the expression

$$t' = t + K_C^{-1}(t \mid t_0) \cdot (F - f(t \mid t_0)) \tag{21}$$

where $t'$ corresponds to the next iteration, $K_C(t \mid t_0)$ is the Cartesian stiffness matrix computed in the point $t$, and $t_0$ denotes the unloaded location of the end-platform. For this iterative scheme, $t_0$ can be also used as the initial value of $t$. Similar to previous sub-section, within each iterative loop, corresponding configurations $(q_i, \theta_i)$, the loadings $F_i$ and stiffness matrices $K_C^{(i)}$ for each kinematic chain are computed using equations (2), and (3) respectively,

As follows from the relevant study, convergence of this iterative procedure is good enough if the function $f(...)$ is smooth and non-singular in the neighborhood of $t_0$. If it is required to improve convergence, it is possible to modify the force $F$ from iteration to iteration in accordance with the expression $F' = \alpha \cdot F$, where a scalar variable $\alpha$ is monotonically increasing from 0 up to 1. The stopping criterion can be expressed in a straightforward way as $\|F - f(t \mid t_0)\| < \varepsilon_F$, where $\varepsilon_F$ is the desired accuracy. A more detailed presentation of the developed iterative routines is given in Fig. 4.

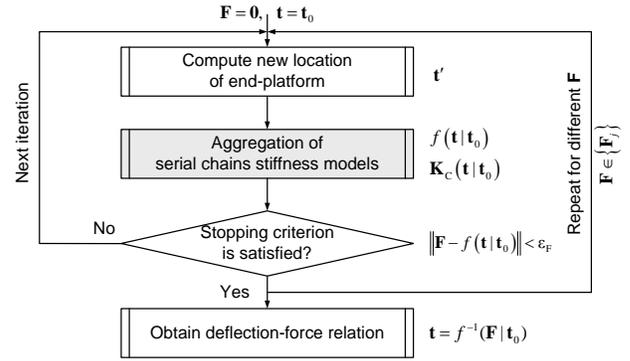

**Fig. 4** Procedure for obtaining deflection-force relation for loaded parallel manipulator

### C. Compliance error compensation technique

To compensate for this undeterred end-platform displacement, the target point should be modified in such a way that, under the loading $F$, the end-platform is located in the desired point $t_0$. This requirement can be expressed using the stiffness model in the following way

$$F = f\left(t_0 \mid t_0^{(F)}\right) \tag{22}$$

where $t_0^{(F)}$ denotes the modified target location. Hence, the problem is reduced to the solution of the nonlinear equation (22) for $t_0^{(F)}$, while $F$ and $t_0$ are assumed to be given. It is worth mentioning that this equation completely differs from the equation $F = f(t \mid t_0)$, where the unknown variable is $t$. It means that here the compliance model does not allow us to compute the modified target point $t_0^{(F)}$ straightforwardly, while the linear compensation technique directly operates with Cartesian compliance matrix.

To solve equation (22) for $t_0^{(F)}$, a similar numerical technique can be applied. It yields the following iterative scheme

$$t_0^{(F)'} = t_0^{(F)} + \alpha \cdot \left(t_0 - f^{-1}(F \mid t_0^{(F)})\right) \tag{23}$$

where the prime corresponds to the next iteration, $\alpha \in (0,1)$ is the scalar parameter ensuring the convergence.

In the case where it is required to compensate for two types of errors (caused by the external loading **F** and inaccuracy in the serial chains), the second source of errors can be taken into account by changing of target location $\Delta \mathbf{t}_{0i}$ for each kinematic chain $\Delta \mathbf{t}_{0i} = \Delta \mathbf{t}_0 + \Delta \mathbf{t}_\varepsilon - \boldsymbol{\varepsilon}_i$, where $\Delta \mathbf{t}_\varepsilon$ is the end-platform deflections due to assembling of non-perfect kinematic chains and $\boldsymbol{\varepsilon}_i$ is shifting of the end-point location of $i^{th}$ kinematic chain because of geometrical errors. More detailed presentation of the developed iterative routines is given in Fig, 5.

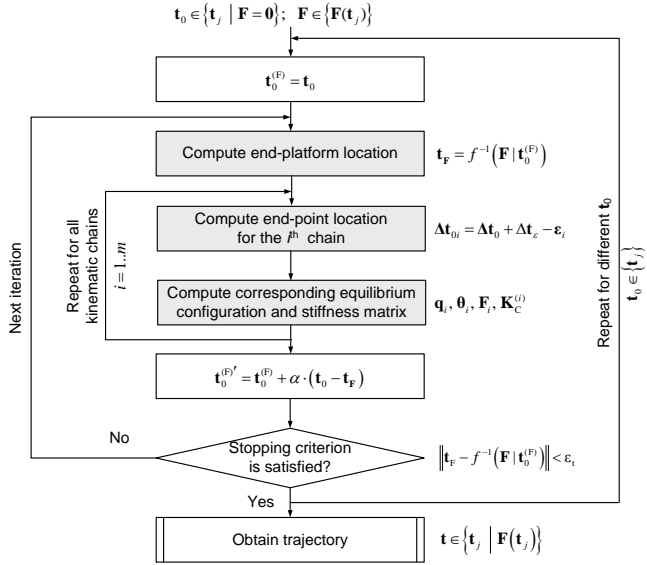

**Fig. 5** Procedure for compensation of compliance errors in parallel manipulator

Hence, using the proposed computational techniques, it is possible to compensate for the main part compliance errors by properly adjusting the reference trajectory that is used as an input for the robotic controller.

## V. APPLICATION EXAMPLES

### A. Aggregation non-perfect serial chains without loading

Let us illustrate the developed stiffness model aggregation technique by examples that deal with assembling of Orthoglide parallel translational manipulators with geometrical errors in kinematic chains (Fig. 6) [23]. Let us assume that the manipulators have geometrical errors in the kinematic chains, which have effects on the end-point location and provoke internal loadings in the joints.

Taking into account the shape of the manipulator workspace, let us focus on the stiffness analysis of these manipulators in five characteristic points: isotropic point $Q_0$, two limit points $Q_1$ and $Q_2$. with symmetrical configuration and two limit points $Q_3$ and $Q_4$. with non-symmetrical configuration [6][7]. Let us estimate the end-effector location and internal deflections/loadings caused by the geometrical errors in the chains. The stiffness matrices of the chains in the points $Q_0...Q_4$ have been computed using the technique proposed in Section III.A.

For illustration purposes, let us investigate two types of geometrical errors

**Case A:** Each actuator of the manipulator has a *position error* 1 mm in actuator location;

**Case B:** Each actuator of the manipulator has an *angular error* 1° in actuator location.

In case A, as it follows from the chains geometry, the deflections of the chain end-points before assembling are $\boldsymbol{\varepsilon}_i = \boldsymbol{\varepsilon}_i^0$. In case B, the values $\boldsymbol{\varepsilon}_i$ were computed using the geometrical model with non-perfect chains:

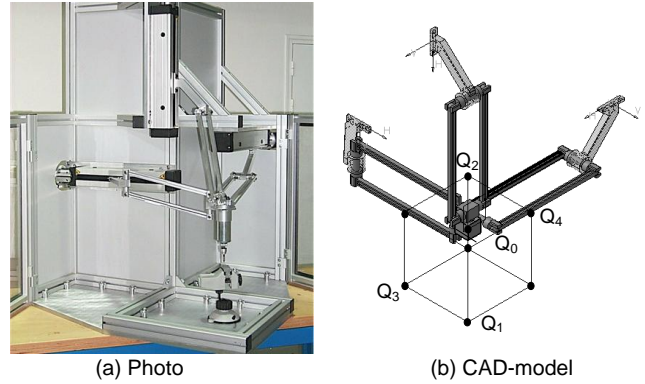

(a) Photo  (b) CAD-model

**Fig. 6** Photo and CAD models of Orthoglide manipulator

Further, substituting deflections $\boldsymbol{\varepsilon}_i$ and corresponding chain stiffness matrices $\mathbf{K}_C^{(i)}$ into formulas (13) - (16), were computed the assembling-induced values of the end-platform displacement, the internal forces/torques and the relevant displacements in virtual and passive joints. The main results of this study are summarized in Tables 1–2, where $\Delta q^{max}$ is the maximum rotational displacement of passive joints, $\theta_p^{max}, \theta_\varphi^{max}$ are maximum displacement of translational and rotational virtual springs respectively, $\tau_p^{max}, \tau_\varphi^{max}$ are maximum torques in translational and rotational virtual joints respectively, $M^{max}$ is the maximum moment in the chains, caused by assembling of a parallel manipulator with the non-perfect kinematic chains.

These results show that in the Case A (Table 1), the geometrical errors in the kinematic chains do not cause any internal loading. However, they provoke the shift of the end-platform location up to 2.02 mm (point $Q_2$). Corresponding changes in passive joint coordinates are about 0.42° (point $Q_2$)

In contrast, in the Case B, the geometrical errors in the kinematic chains of Orthoglide (Table 2) cause essential internal loadings. For instance, in point $Q_1$ the torque applied to the end-point of the chain can reach up to $8.91 \, \text{N} \cdot \text{m}$. This loading induces displacements up to $0.41 \, \text{mm}$ and $0.62°$ for translational and rotational virtual springs respectively. It should be noted that the loadings for rotational virtual springs are up to $11.96 \, \text{N} \cdot \text{m}$, but for translational virtual springs they are equal to zero (in spite of non-zero deflections in them). Nevertheless, this result is reasonable due to the non-diagonal structure of the matrices $\mathbf{K}_C^{(i)}$ representing couplings between rotational and translational deflections. Variations in the passive joint coordinates can reach up to $0.67°$ (Point $Q_3$). For the end-platform, this case study gives the positional deflection up to 1.31 mm (Point $Q_3$) and the rotational deflection up to $0.62°$ (Point $Q_1$). It is obvious that the total sum of all internal loadings is equal to zero

**Table 1.** Assembling of Orthoglide manipulator with non-perfect chains: loadings and displacements for the Case A ($\Delta \mathbf{t} = [\delta_1, \delta_2, \delta_3, 0, 0, 0]^T$, $\mathbf{F}_1 = \mathbf{0}$, $\mathbf{F}_2 = \mathbf{0}$, $\mathbf{F}_3 = \mathbf{0}$)

| Point | Displacement of end-point $\Delta \mathbf{t}$ | Deflections and loadings in joints and links |
|---|---|---|
| $Q_0$ | $\delta_1 = \delta_2 = \delta_3 = 1$ mm; | $\Delta q^{max} = 0.18°$ |
| $Q_1$ | $\delta_1 = \delta_2 = \delta_3 = 0.50$ mm; | $\Delta q^{max} = 0.14°$ |
| $Q_2$ | $\delta_1 = \delta_2 = \delta_3 = 2.02$ mm; | $\Delta q^{max} = 0.42°$ |
| $Q_3$ | $\delta_1 = \delta_2 = 0.73$ mm; $\delta_3 = 0.27$ mm | $\Delta q^{max} = 0.20°$ |
| $Q_4$ | $\delta_1 = \delta_2 = 0.56$ mm; $\delta_3 = 1.28$ mm | $\Delta q^{max} = 0.26°$ |
| | $\theta_p^{max} = 0$; $\theta_\varphi^{max} = 0$; $\tau_p^{max} = 0$; $\tau_\varphi^{max} = 0$ | |

**Table 2.** Assembling of Orthoglide manipulator with non-perfect chains: loadings and displacements for the case B ($\Delta \mathbf{t} = [\delta_1, \delta_2, \delta_3, \varphi_1, \varphi_2, \varphi_3]^T$, $\mathbf{F}_1 \ne \mathbf{0}$, $\mathbf{F}_2 \ne \mathbf{0}$, $\mathbf{F}_3 \ne \mathbf{0}$)

| Point | Displacement of end-point $\Delta \mathbf{t}$ | Deflections and loadings in joints and links |
|---|---|---|
| $Q_0$ | $\delta_1 = \delta_2 = \delta_3 = 0$ mm; $\varphi_1 = \varphi_2 = \varphi_3 = 0.03°$; | $M^{max} = 2.09$ N·m; $\Delta q^{max} = 0.31°$ $\theta_p^{max} = 0.05$ mm; $\theta_\varphi^{max} = 0.94°$ $\tau_p^{max} = 0$; $\tau_\varphi^{max} = 2.09$ N·m |
| $Q_1$ | $\delta_1 = \delta_2 = \delta_3 = 0.41$ mm; $\varphi_1 = \varphi_2 = \varphi_3 = -0.62°$; | $M^{max} = 8.91$ N·m; $\Delta q^{max} = 0.63°$ $\theta_p^{max} = 0.54$ mm; $\theta_\varphi^{max} = 1.74°$ $\tau_p^{max} = 0$; $\tau_\varphi^{max} = 11.96$ N·m |
| $Q_2$ | $\delta_1 = \delta_2 = \delta_3 = -0.96$ mm; $\varphi_1 = \varphi_2 = \varphi_3 = 0.21°$; | $M^{max} = 1.48$ N·m; $\Delta q^{max} = 0.52°$ $\theta_p^{max} = 0.14$ mm; $\theta_\varphi^{max} = 0.80°$ $\tau_p^{max} = 0$; $\tau_\varphi^{max} = 1.75$ N·m |
| $Q_3$ | $\delta_1 = -0.91$ mm; $\varphi_1 = -0.19°$ $\delta_2 = 1.31$ mm; $\varphi_2 = -0.49°$ $\delta_3 = 0.58$ mm; $\varphi_3 = 0.44°$ | $M^{max} = 4.33$ N·m; $\Delta q^{max} = 0.67°$ $\theta_p^{max} = 0.99$ mm; $\theta_\varphi^{max} = 1.49°$ $\tau_p^{max} = 0$; $\tau_\varphi^{max} = 4.84$ N·m |
| $Q_4$ | $\delta_1 = 0.93$ mm; $\varphi_1 = 0.33°$ $\delta_2 = -0.10$ mm; $\varphi_2 = 0.22°$ $\delta_3 = -0.25$ mm; $\varphi_3 = -0.31°$ | $M^{max} = 2.98$ N·m; $\Delta q^{max} = 0.59°$ $\theta_p^{max} = 0.62$ mm; $\theta_\varphi^{max} = 1.30°$ $\tau_p^{max} = 0$; $\tau_\varphi^{max} = 4.0$ N·m |

### B. Aggregation non-perfect serial chains under loading

Now let us consider the chain stiffness model aggregation of Orthoglide manipulator under external loading caused by groove milling.. According to [24], such technological process causes forces $F_r = 215$ N; $F_t = -10$ N; $F_z = -25$ N. The tool length $h = 100$ mm leads to torques at the manipulator end-effector $M_x = 1$ N·m and $M_y = 21.5$ N·m. It is assumed that the manipulator has two sources of inaccuracy:

(i) the assembling errors in the kinematic chains causing internal forces and relevant deflections in joints and links due to manipulator over-constrained structure;
(ii) the external loading $\|\mathbf{F}\| = 217$ N caused by the cutting force, which generates essential compliance deflections causing non-desirable end-platform displacement.

Similar to the previous example, it is assumed that the first source of inaccuracy can be caused by translational (Case A) and rotational (Case B) errors in the actuator locations.

Let us consider the case study when Orthoglide performs milling from the point $Q_2$ to $Q_5$(-73.65, 126.35, -73.65) following the straight line. Simulation results for two error sources for the Case A and Case B are presented in Fig 7 and Fig. 8 respectively. They include the target trajectories, displacements caused by the cutting forces and non-perfect geometry as well as total compliance errors and the displacement evaluated using the superposition principle. The results are presented for the displacements in x- and z-directions.

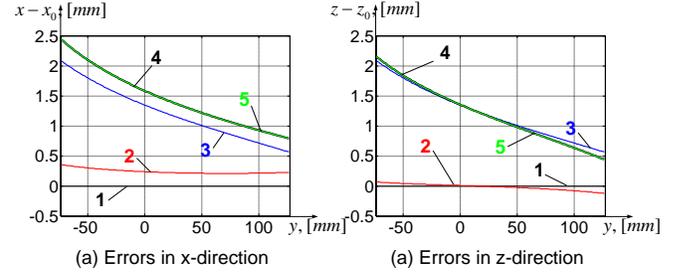

**Fig. 7.** Case A: Displacements caused different sources of inaccuracy during milling along the straight line from point $Q_2$ to $Q_4$ using Orthoglide manipulator: (1) target trajectory, (2) displacements caused by cutting forces, (3) displacements caused by non-perfect geometry, (4) total compliance error, (5) displacements obtained using superposition principle.

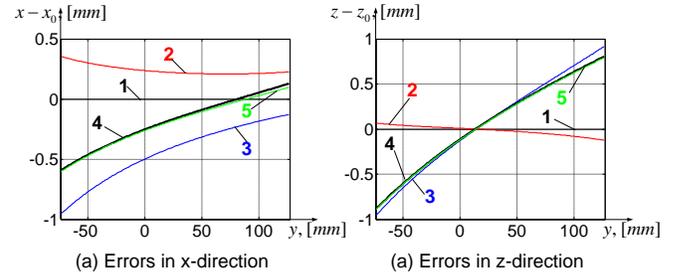

**Fig. 8.** Case B: Displacements caused different sources of inaccuracy during milling along the straight line from point $Q_2$ to $Q_4$ using Orthoglide manipulator: (1) target trajectory, (2) displacements caused by cutting forces, (3) displacements caused by non-perfect geometry, (4) total compliance error, (5) displacements obtained using superposition principle.

The obtained results show that, due to the cutting force, the end-platform displacement in the x-direction with respect to the target trajectory may reach up to 0.4 mm and its variations are insignificant (compared to the errors caused by inaccuracy in the serial chains). In particular, in Case A, the errors in the x-direction induced by cutting forces and inaccuracy in serial chains have the same direction, and, consequently total error is close to 2.5 mm in the point $Q_2$ and about 0.8 mm in the vicinity of the point $Q_4$. In contrast, in the Case B, the compliance errors in the x-direction induced by the cutting forces and inaccuracy in the serial chains have different signs, so the total error in the loaded mode is less than in the unloaded one. This error varies from -0.6 mm to 0.1 mm along the trajectory and even reduced to zero when the y-coordinate is about 80 mm. In the z-direction, the errors caused by the assembling of non-perfect serial chains is much higher (from 0.5 mm to 2.1 mm for the Case A and from -1 mm to 1 mm in the Case B). To compensate for these errors the technique presented in sub-section IV.C can be applied.

## VI. CONCLUSIONS

This paper presents the non-linear stiffness modeling technique for parallel manipulators composed of non-perfect serial chains, whose geometry differ from the nominal one and where essential internal forces/torques are generated. This technique is based on the developed aggregation procedure that combines the chain stiffness models and produces the relevant force-deflection relation, the aggregated Cartesian stiffness matrix and also allows us to evaluate changes in the reference point location caused by inaccuracy in kinematic chains. In addition, expressions for computing of the internal deflections and forces/torques in joints are proposed. The developed technique can be applied to both over-constrained and under-constrained manipulators, and is suitable for the cases of both small and large deflections.

The advantages of the developed technique are illustrated by an example that deals with the over-constrained parallel manipulator of the Orthoglide architecture. It demonstrates the technique ability to evaluate the end-effector deflections caused by conventional sources (cutting forces/torques applied to the end-effector that arise while workpiece processing) and also induced by inaccuracy in serial chains of the parallel manipulator. Relevant plots that illustrate influence of different error sources on the manipulator position accuracy are presented.

In future, the proposed technique will be integrated in a software toolbox that can be used for parallel manipulators of complex architecture and applied to the industrial problem of the compliance error compensation in robotic machining cells.

ACKNOWLEDGEMENTS

The work presented in this paper was partially funded by the Region "Pays de la Loire" (France), and by the ANR, France (project COROUSSO).

REFERENCES

[1] O. Company, S. Krut, F. Pierrot, Modelling and preliminary design issues of a 4-axis parallel machine for heavy parts handling, Journal of Multibody Dynamics 216 (2002) 1–11.
[2] J. Kövecses, J. Angeles, The stiffness matrix in elastically articulated rigid-body systems, Multibody System Dynamics 18(2) (2007) 169–184.
[3] T. Bonnemains, H. Chanel, C. Bouzgarrou and P. Ray, Definition of a new static model of parallel kinematic machines: highlighting of overconstraint influence, in: Proceedings of IEEE Int. Conference on Intelligent Robots and Systems (IROS), 2008, pp. 2416–2421.
[4] D. Deblaise, X. Hernot, P.Maurine, A systematic analytical method for PKM stiffness matrix calculation, In: Proceedings of the IEEE International Conference on Robotics and Automation (ICRA), Orlando, Florida, 2006, pp. 4213-4219.
[5] Y. Li, Q. Xu, Stiffness analysis for a 3-PUU parallel kinematic machine, Mechanism and Machine Theory 43(2) (2008) 186-200.
[6] A. Pashkevich, D. Chablat, P. Wenger, Stiffness analysis of overconstrained parallel manipulators, Mechanism and Machine Theory 44 (2009) 966-982.
[7] A. Pashkevich, A. Klimchik, D. Chablat, "Enhanced stiffness modeling of manipulators with passive joints", Mechanism and Machine Theory, vol. 46(5), pp. 662-679, 2011.
[8] A. Pashkevich, A. Klimchik, S. Caro, D. Chablat, Cartesian stiffness matrix of manipulators with passive joints: analytical approach, In: IEEE/RSJ International Conference on Intelligent Robots and Systems (IROS 2011), September 25-30, 2011, USA, San Francisco, California, pp. 4034-4041.
[9] J. Salisbury, Active Stiffness Control of a Manipulator in Cartesian Coordinates, in: 19th IEEE Conference on Decision and Control, 1980, pp. 87–97.
[10] C. Gosselin, Stiffness mapping for parallel manipulators, IEEE Transactions on Robotics and Automation 6(3) (1990) 377–382.
[11] S. Chen, I. Kao, Conservative Congruence Transformation for Joint and Cartesian Stiffness Matrices of Robotic Hands and Fingers, The International Journal of Robotics Research 19(9) (2000) 835–847
[12] C. Quennouelle, C. M. Gosselin, Stiffness Matrix of Compliant Parallel Mechanisms, In: Springer Advances in Robot Kinematics: Analysis and Design, 2008, pp. 331-341.
[13] I. Tyapin, G. Hovland, Kinematic and elastostatic design optimization of the 3-DOF Gantry-Tau parallel kinamatic manipulator,Modelling, Identification and Control, 30(2) (2009) 39-56
[14] G. Alici, B. Shirinzadeh, Enhanced stiffness modeling, identification and characterization for robot manipulators, Proceedings of IEEE Transactions on Robotics 21(4) (2005) 554–564.
[15] A. Pashkevich, A. Klimchik, D. Chablat, Ph. Wenger, Accuracy Improvement for Stiffness Modeling of Parallel Manipulators, In: Proceedings of 42nd CIRP Conference on Manufacturing Systems, Grenoble, France, 2009.
[16] J.-P. Merlet, Parallel Robots, Kluwer Academic Publishers, Dordrecht, 2006.
[17] D. Zhang, F. Xi, C.M. Mechefske, S.Y.T. Lang, Analysis of parallel kinematic machine with kinetostatic modeling method, Robotics and Computer-Integrated Manufacturing 20 (2) (2004) 151–165
[18] D. Chakarov, "Study of antagonistic stiffness of parallel manipulators with actuation redundancy," Mechanism and Machine Theory. Vol. 39(6), pp. 583–601, 2004.
[19] B.-J. Yi, R.A. Freeman, "Geometric analysis antagonistic stiffness redundantly actuated parallel mechanism," Journal of Robotic Systems 10(5) (1993) 581-603.
[20] F. Xi, D. Zhang, Ch. M. Mechefske, Sh. Y. T. Lang, "Global kinetostatic modelling of tripod-based parallel kinematic machine," Mechanism and Machine Theory 39 (4), pp. 357–377, 2004.
[21] R. Rizk, N. Andreff, J.C. Fauroux, J.M. Lavest and G. Gogu, Precision Study of a Decoupled Four Degrees of Freedom Parallel Robot Including Manufacturing and Assembling Errors, Advances in Integrated Design and Manufacturing in Mechanical, S. Tichkiewitch et al. (eds.), Springer 2007, Engineering II, 111–127.
[22] W. Wei, N. Simaan, "Design of planar parallel robots with preloaded flexures for guaranteed backlash prevention," ASME Journal of Mechanisms and Robotics, Vol. 2(1) , 10 pages, 2010
[23] D. Chablat, P. Wenger, Architecture Optimization of a 3-DOF Parallel Mechanism for Machining Applications, the Orthoglide, IEEE Transactions On Robotics and Automation 19(3) (2003) 403-410.
[24] F. Majou, C. Gosselin, P. Wenger, D. Chablat, Parametric stiffness analysis of the Orthoglide, Mechanism and Machine Theory 42 (2007) 296-311.